\documentclass[letterpaper, 10 pt, conference]{ieeeconf}  % Comment this line out if you need a4paper

\IEEEoverridecommandlockouts                              % This command is only needed if 
\usepackage{graphics} % for pdf, bitmapped graphics files
\usepackage{epsfig} % for postscript graphics files
\usepackage{mathptmx} % assumes new font selection scheme installed
\usepackage{times} % assumes new font selection scheme installed
\usepackage{amsmath} % assumes amsmath package installed
\usepackage{amssymb}  % assumes amsmath package installed
\usepackage{bm}

\PassOptionsToPackage{hyphens}{url}\usepackage{hyperref}
\usepackage[caption=false]{subfig} % Subfigures

\usepackage{xcolor}

\usepackage[style=ieee]{biblatex}

\DeclareSourcemap{
  \maps{
    \map{
      \pertype{article}
      \step[fieldset=language, null]
      \step[fieldset=url, null]
      \step[fieldset=doi, null]
      \step[fieldset=issn, null]
      \step[fieldset=isbn, null]
      \step[fieldset=note, null]
      \step[fieldset=editor, null]
      \step[fieldset=urldate, null]
      \step[fieldset=file, null]
    }
  }
}
\DeclareSourcemap{
  \maps{
    \map{
      \pertype{inproceedings}
      \step[fieldset=language, null]
      \step[fieldset=url, null]
      \step[fieldset=doi, null]
      \step[fieldset=issn, null]
      \step[fieldset=isbn, null]
      \step[fieldset=note, null]
      \step[fieldset=editor, null]
      \step[fieldset=urldate, null]
      \step[fieldset=file, null]
    }
  }
}
\DeclareSourcemap{
  \maps{
    \map{
      \pertype{incollection}
      \step[fieldset=language, null]
      \step[fieldset=url, null]
      \step[fieldset=doi, null]
      \step[fieldset=issn, null]
      \step[fieldset=isbn, null]
      \step[fieldset=note, null]
      \step[fieldset=editor, null]
      \step[fieldset=urldate, null]
      \step[fieldset=file, null]
    }
  }
}

\title{\LARGE \bf
Modeling and Controls of Fluid-Structure Interactions (FSI) in Dynamic Morphing Flight
}

\author{Bibek Gupta$^{1}$, Eric Sihite$^{2}$ and Alireza Ramezani$^{1*}$% <-this % stops a space
\thanks{$^{1}$Authors are with the Silicon Synapse Labs, Department of
Electrical and Computer Engineering, Northeastern University, Boston,
USA. Emails: gupta.bi, a.ramezani@northeastern.edu}%
\thanks{$^{2}$Author is with the Department of Aerospace Engineering,
California Institute of Technology, Pasadena, USA. Email:
esihite@caltech.edu}%
\thanks{*Corresponding author.}% <-this % stops a space
}

\begin{document}

\maketitle
\thispagestyle{empty}
\pagestyle{empty}

\begin{abstract}
The primary aim of this study is to enhance the accuracy of our aerodynamic Fluid-Structure Interaction (FSI) model to support the controlled tracking of 3D flight trajectories by Aerobat, which is a dynamic morphing winged drone. Building upon our previously documented Unsteady Aerodynamic model rooted in horseshoe vortices, we introduce a new iteration of Aerobat, labeled as version $\beta$, which is designed for attachment to a Kinova arm. Through a series of experiments, we gather force-moment data from the robotic arm attachment and utilize it to fine-tune our unsteady model for banking turn maneuvers. Subsequently, we employ the tuned FSI model alongside a collocation control strategy to accomplish 3D banking turns of Aerobat within simulation environments. The primary contribution lies in presenting a methodical approach to calibrate our FSI model to predict complex 3D maneuvers and successfully assessing the model's potential for closed-loop flight control of Aerobat using an optimization-based collocation method.
\end{abstract}

\section{Introduction}
\label{sec:intro}

Dynamic morphing winged systems, akin to bats or birds, dexterously manipulate their fluidic environment \cite{hubel_wake_2010}. This ability to modulate fluid momentum through controlled movements in three dimensions distinguishes vertebrate flight from that of insects \cite{karpelson_review_2008} and is key to vertebrates' agile and efficient flight.

Research on drones with adaptive body structures similar to bats and birds has been ongoing for many years \cite{bae_aerodynamic_2005}. These systems often possess slow moving joints that do not match dynamic morphing capabilities seen in vertebrate fliers. For instance, bats are able to move over forty joints dynamically during one gait cycle \cite{riskin_upstroke_2012}. 

The emergence of small form factor electronics and actuators has enabled the development of new small robots capable of dynamically adjusting their body configurations dynamically \cite{chang_soft_2020,di_luca_bioinspired_2017,vincent_flying_2017,hoff_optimizing_2018}. The Northeastern University Aerobat \cite{sihite_computational_2020,sihite_unsteady_2022,ramezani_lagrangian_2015,ramezani_bat_2016,sihite_computational_2020,hoff_reducing_2017,ramezani_describing_2017,hoff_synergistic_2016,sihite_wake-based_2022} aims to perfect controlled fluidic manipulations. This robot has an articulated wing structure as is able to dynamically move its joints during one wingcycle, about one tenth of a second.

\begin{figure}
    \centering
    \includegraphics[width=0.9\linewidth]{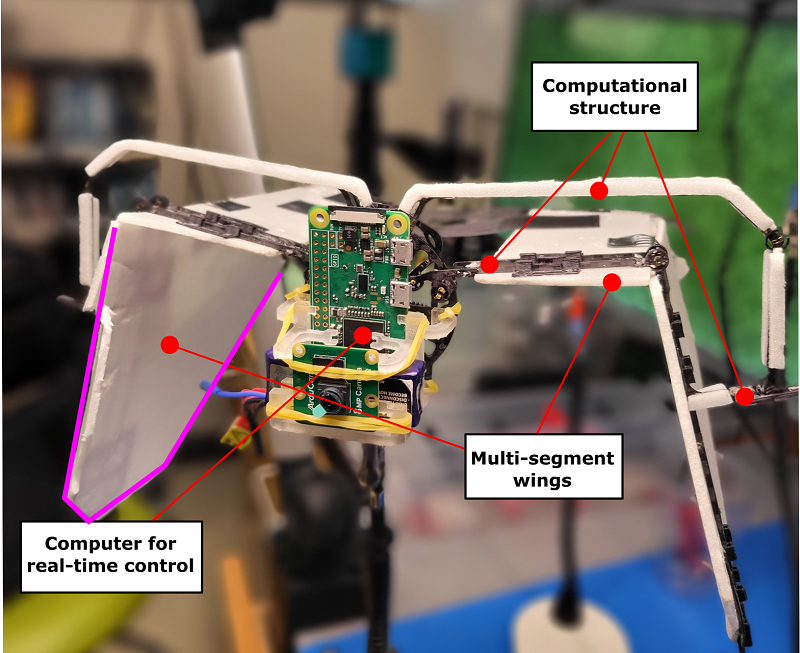}
    \caption{Shows Northeastern University Aerobat platform designed to inspect dynamic morphing wing flight}
    \label{fig:cover-image}
\end{figure}

Dynamic morphing of wings introduces complex fluid-structure interactions, rendering existing quasi-steady models used in insect flight inadequate \cite{lee_quasi-steady_2016-1}. Moreover, these morphing wing robots feature heavier wings that deviate from the two-time scale dynamical models typically applied to insects \cite{schenato_controllability_2003}. Consequently, these systems not only present distinctive opportunities for hardware design but also offer rich prospects for modeling and control.

In our previous studies, we introduced unsteady fluid-structure models predicting the force-moment profile in Aerobat flight \cite{sihite_unsteady_2022} motivated by prior works such as \cite{boutet_unsteady_2018} and \cite{izraelevitz_state-space_2017}. This model accommodated the step response of the lift coefficient as the morphing wing undergoes plunge movements and flexion-extensions in the armwings. 

We experimentally validated this model's accuracy for hovering maneuvers by subjecting Aerobat to varying upstream flow conditions \cite{dhole_hovering_2023}. Based on these results, we applied this model to design a hovering controller in a tandem configuration in \cite{dhole_hovering_2023}. The unsteady model, coupled with an extended state observer (ESO), estimated FSI in the tandem structure, which was then used to control the tandem platform.

In this study, our overarching objective is to expand Aerobat's flight capabilities beyond hovering, encompassing banking turns, sharp dives, heel-above-the-head landings, and more. These locomotive maneuvers demand enhanced accuracy of the unsteady model as Aerobat traverses complex 3D paths. The main contributions of this publication are: i) the presentation of the test platform developed for tuning the unsteady model as Aerobat navigates complex 3D paths, ii) the collection and comparison of experimental data with the unsteady simulator, and iii) the utilization of the tuned model to design a collocation-based controller for tracking a 3D path in simulation. 

This work is organized as follows: we begin with the hardware and constrained model overview, followed by the summary of the unsteady aerodynamic model used to estimate the fluid-structure interactions (FSI), 3D path tracking and control. Then, it will be followed by the result discussions and concluding remarks.

\section{Constrained Model Hardware Overview}
\label{sec:hdw}

\begin{figure*}
    \centering
    \includegraphics[width=\textwidth]{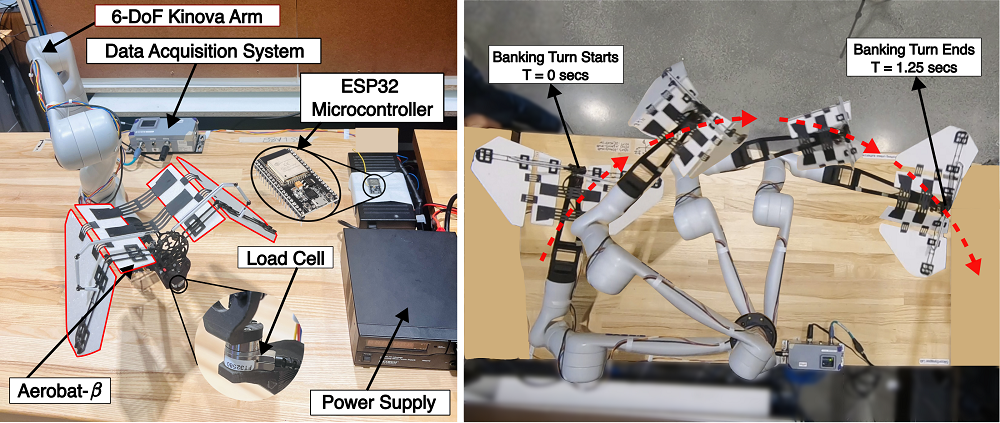}
    \caption{(a) Shows the test setup, including the arm, load-cell, Aerobat version $\beta$, and data acquisition system. (b) Snapshots of arm and Aerobat during the constrained banking turn at different sample times are overlaid and illustrated in this image.}
    \label{fig:test-setup}
\end{figure*}

To emulate 3D path flights and record FSI from experiments, we design a version of Aerobat, called version-$\beta$, and fixated the design to the end-effector of a Kinova arm as shown in Fig.~\ref{fig:test-setup}. Aerobat-$\beta$, similar to other versions \cite{sihite_unsteady_2022}, possesses a computational structure \cite{sihite_mechanism_2020} that generates dynamic coupled joint movements. The resulting wing movements include (1) plunge movements at the shoulders and (2) flexion-extension motions at the elbows. A gait generator, a brushless-dc motor with gearbox, drives the computational structure.

The Aerobat-$\beta$ model hardware lacks orientation control, rendering it incapable of executing banking turns independently in untethered fashion. To overcome this limitation, for model turning purposes, this study utilizes a Gen-3 6-DOF Kinova robotic arm (see Fig.~\ref{fig:test-setup}), characterized by its joint speed reaching up to 70 degrees per second and a linear end-effector speed capacity of up to 50 cm/s.

\begin{figure*}[t]
    \centering
    \includegraphics[width=0.9\textwidth]{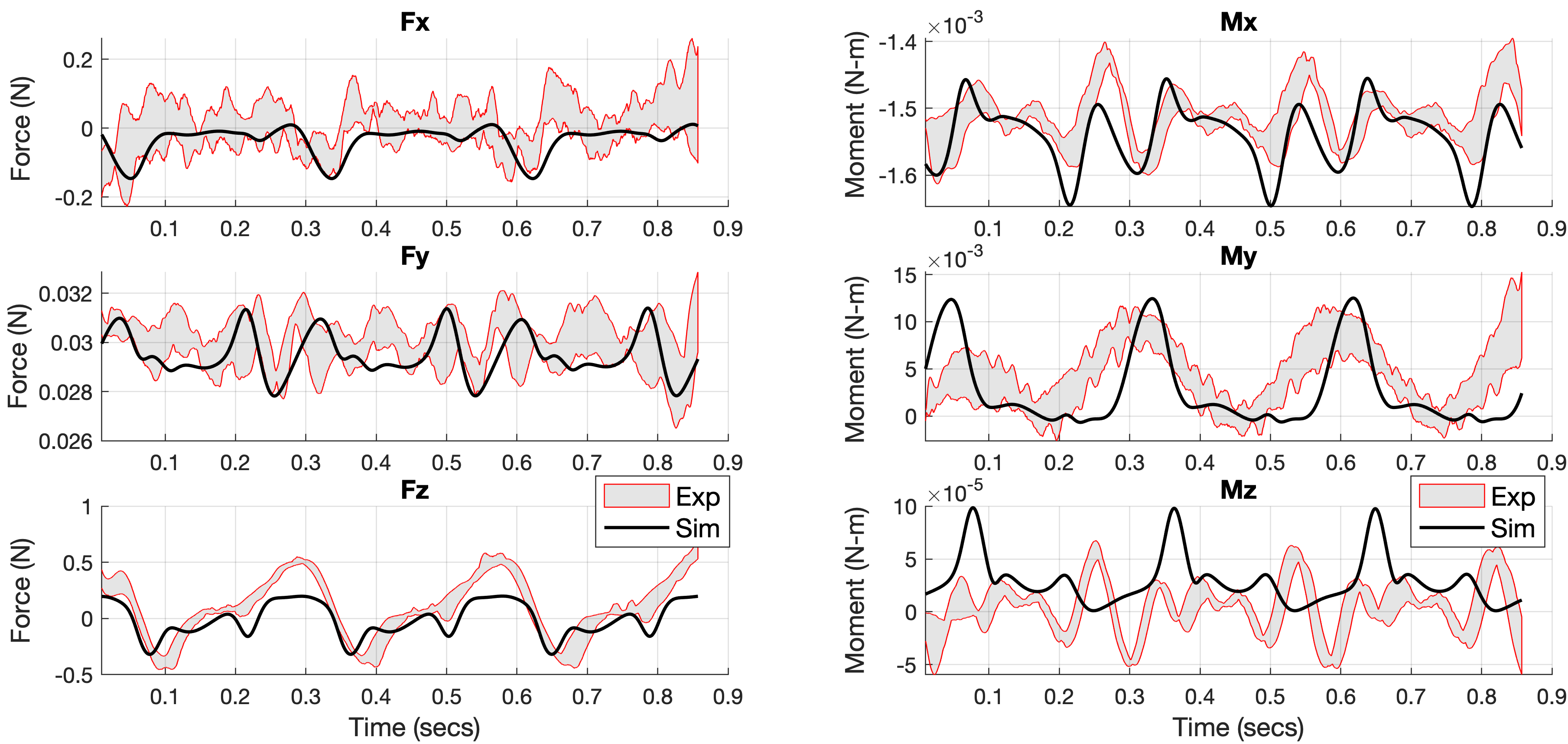}
    \caption{Illustrates comparison of the force-moment trajectories between experiment and simulation.}
    \label{fig:force-moment}
\end{figure*}

Next, as shown in Fig.~\ref{fig:test-setup}, through the integration of the Aerobat robot onto the Kinova arm's end-effector via a custom-designed 3D printed mount equipped with an ATI load cell, we recorded force-moment profile during the execution of banking turns as shown in Fig.~\ref{fig:force-moment}. This ATI load cell is a Nano17 6-axis-transducer model, having a Net box for data acquisition via Ethernet, and has rated capacities of 480 N for load force and 1.9 N·m for load torque. Using the load cell, data for both force and moments were collected at a sampling frequency of 7 kHz during the banking maneuver experiments.

\section{Unsteady Model Used to Predict Fluid-Structure Interactions (FSI)}
\label{sec:fsi}

% \begin{figure}
%     \centering
%     \includegraphics[width=0.9\linewidth]{example-image-a}
%     \caption{Illustrates model parameters used to obtain force-moment obtained in Section~\ref{sec:fsi}}
%     \label{fig:enter-label}
% \end{figure}

\begin{figure*}[t]
    \centering
    \includegraphics[width=0.9\textwidth]{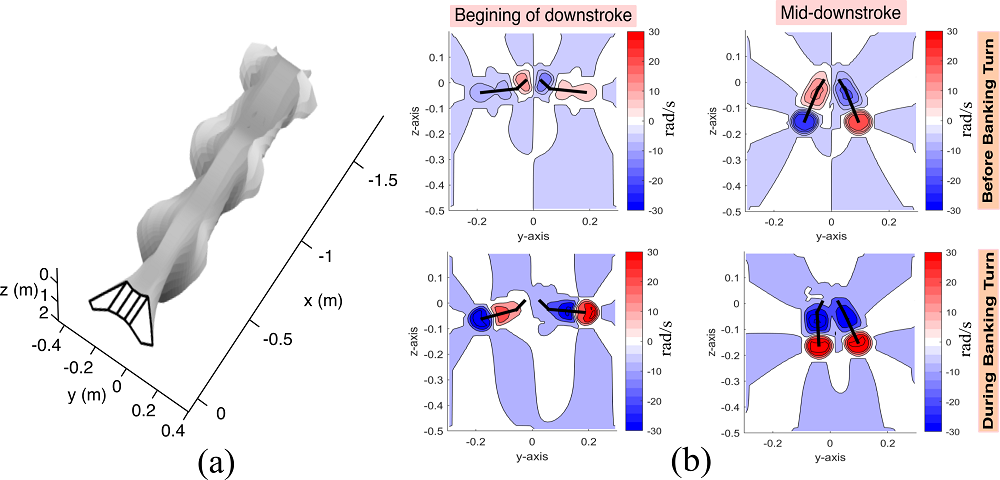}
    \caption{Illustrates wake structures (iso-metric view) and vorticity plots in the frontal plane of flapping at (i) the beginning of downstroke, (ii) middle of downstroke, (iii) beginning of upstroke, and (iv) the middle of upstroke during banking turn.}
    \label{fig:wakes}
\end{figure*}

\begin{figure*}[t]
    \centering
    \includegraphics[width=0.9\textwidth]{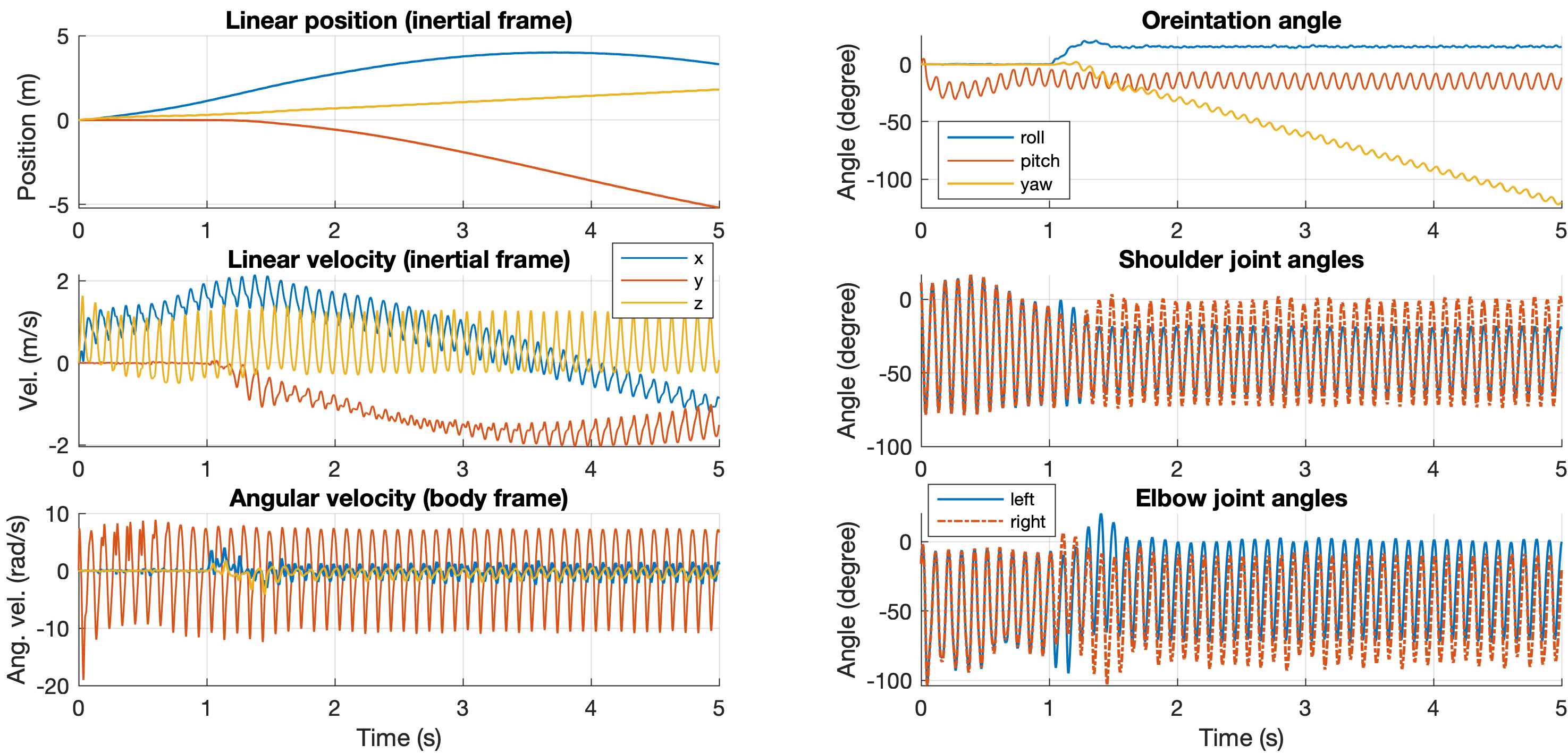}
    \caption{Shows all of Aerobat state trajectories including body orientation, position, and body angles (shoulder and elbow joints).}
    \label{fig:states}
\end{figure*}

\begin{figure}[t]
    \centering
    \includegraphics[width=0.9\linewidth]{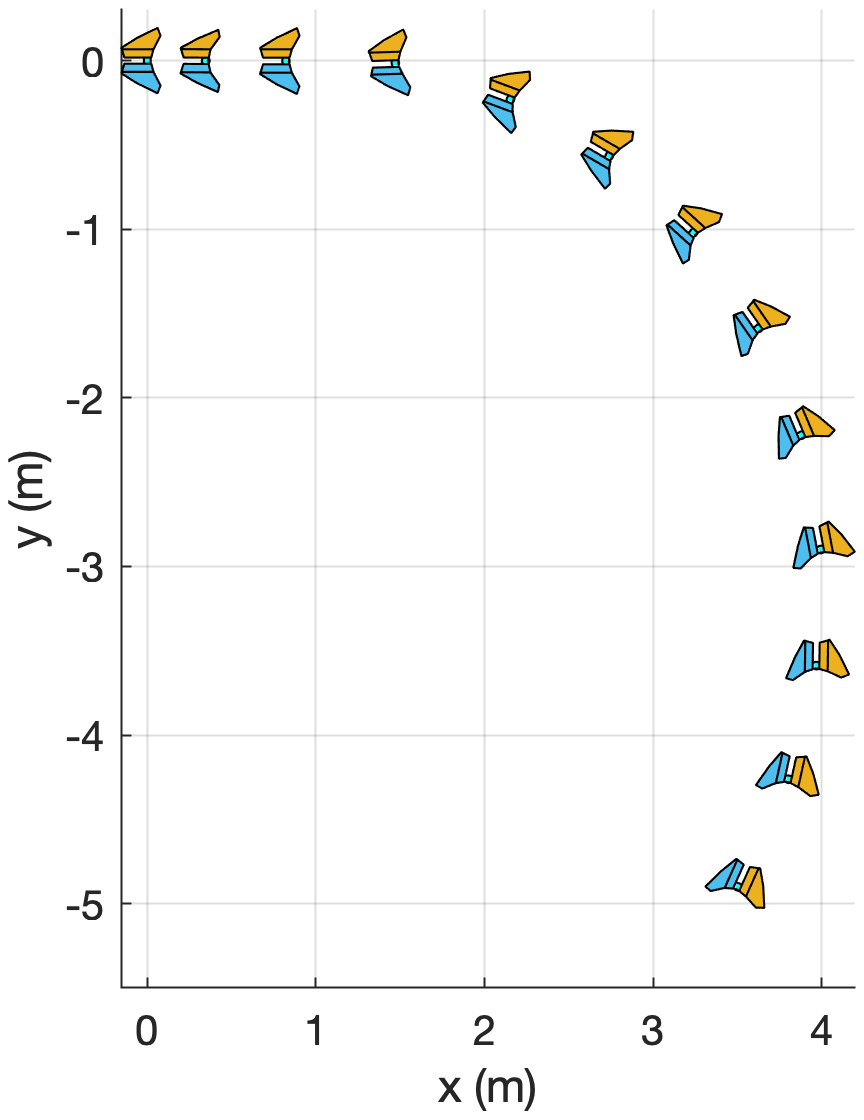}
    \caption{Illustrates snapshots of Aerobat performing controlled 3D path tracking (banking turn) by utilizing a collocation controller to regulate its actuators.}
    \label{fig:sim-stick-diagram}
\end{figure}

The unsteady lift aerodynamic model derived in this section follows similar derivations to \cite{boutet_unsteady_2018}. This model uses the lifting line theory and Wagner's function to develop a model for calculating the lift coefficient. Let $S$ be the total wingspan and $y \in [-S/2, S/2]$ represents a position along the wingspan. The circulation distribution on the wing can be defined as a function of truncated Fourier series of size $m$ across the wingspan, as follows:
\begin{equation}
\begin{gathered}
    \Gamma(t,y) = \frac{1}{2} a_0 \, c_0 \, U \, \sum^{m}_{n=1} a_n(t) \, \sin(n\,\theta(y))
\end{gathered}
\end{equation}
where $a_n$ is the Fourier coefficients, $a_0$ is the slope of the angle of attack, $c_0$ is the chord length at wing's axis of symmetry, and $U$ is the free stream airspeed. Then $\theta$ is the change of variable defined by $y = (S/2)\cos(\theta)$ for describing a position along the wingspan $y \in (-S/2, S/2)$. From $\Gamma(t,y)$, we can derive the additional downwash induced by the vortices, defined as follows:
\begin{equation}
\begin{aligned}
    w_{y}(t,y) & 
    = -\frac{1}{4\pi} \int_{-S/2}^{S/2} \frac{d \Gamma / d y_0}{y - y_0} dy_0 \\ &
    = - \frac{a_0 c_0 U}{4S} \sum^{m}_{n=1} n a_n(t)  \frac{\sin(n \theta)}{\sin(\theta)}.
\end{aligned}
\label{eq:induced_downwash}
\end{equation}
%
%Note that \eqref{eq:induced_downwash} has a division by $\sin(\theta)$, which means that this equation is singular for $\theta \in \{0, \pm \pi\}$, or the $\theta$ values representing the wingtip positions.

Following the unsteady Kutta-Joukowski theorem, the sectional lift coefficient can be expressed as follows:
\begin{equation}
\begin{aligned}
    C_L(t,y) &= \frac{2 \Gamma}{U c(y)} + \frac{2\dot{\Gamma}}{U^2} \\
        &= a_0 \sum^{m}_{n=1} \left( \frac{c_0}{c(y)} a_n(t) + \frac{c_0}{U} \dot{a}_n(t) \right) \sin(n\theta),
\end{aligned}
\label{eq:lift_coeff_fourier}
\end{equation}
where $c(y)$ is the chord length at the wingspan position $y$. The computation of the sectional lift coefficient response of an airfoil undergoing a step change in downwash $\Delta w(y) \ll U$ can be expressed using Wagner function $\Phi(t)$:
\begin{equation}
\begin{aligned}
    c_L(t,y) &= \frac{a_0}{U} \Delta w(t,y) \Phi(\tilde t) \\
    \Phi(\tilde t)    &= 1 - \psi_1 e^{-\epsilon_1 \tilde t} - \psi_2 e^{-\epsilon_2 \tilde t}
\end{aligned}
\label{eq:lift_coeff_wagner}
\end{equation}
where $\tilde t(t) = \int_0^t (v_e^i/b) dt$ is the normalized time which is defined as the distance traveled divided by half chord length ($b = c/2$). Here, $v_e^i$ is defined as the velocity of the quarter chord distance from the leading edge in the direction perpendicular to the wing sweep. For the condition where the freestream airflow dominates $v_e$, then we can approximate the normalized time as $\tilde t = Ut/b$. The Wagner model in \eqref{eq:lift_coeff_wagner} uses Jones' approximation \cite{boutet_unsteady_2018}, with the following coefficients: $\psi_1 = 0.165$, $\psi_2 = 0.335$, $\epsilon_1 = 0.0455$, and $\epsilon_2 = 0.3$.

\section{3D Path Tracking and Controls}
\label{sec:ctrl}

To solve this flight control problem, i.e., Aerobat's posture $y_1$ is recruited to regulate fluid-structure forces-moments $y_2$ to track a 3D path, we consider the following cost function given by
\begin{equation}
    J = \sum_i^N (z_{i} - z_{ref,i})^\top \, C \, (z_{i} - z_{ref,i}),
    \quad
    z = [\theta_r, \theta_p, \omega^\top]^\top,
    \label{eq:cost}
\end{equation}
where $\theta_{r}$ and $\theta_{p}$ represent the robot's roll and pitch angles relative to the inertial frame, respectively, while $\omega$ represents the robot's angular velocities, $z$ is the optimization state, $z_{ref}$ is the state reference for $z$, and $C$ is a diagonal cost weighting matrix. The cost function $J$ is governed by a system of $n$ nonlinear equations representing the computational structure dynamics driven by low-power actuators.
% , as depicted in Fig.~\ref{fig:ks}.

To further elucidate, following the principle of virtual work \cite{landau1982mechanics}, the response from the computational structure can be determined by
\begin{equation}
\begin{aligned}
    \begin{bmatrix}
    \dot y_{1,1}\\
    \ddot y_{1,1}\\
    \vdots\\
    \dot y_{1,n}\\
    \ddot y_{1,n}\\
    \end{bmatrix}=&
    \begin{bmatrix} 
    a_{1,1} & a_{1,2} & \dots \\
    \vdots & \ddots & \\
    a_{2n,1} &        & a_{2n,2n} 
    \end{bmatrix}
    \begin{bmatrix}
    y_{1,1}\\
    \dot y_{1,1}\\
    \vdots\\
    y_{1,n}\\
    \dot y_{1,n}\\
    \end{bmatrix}
    +\\
    & \hspace{1cm} \begin{bmatrix} 
    b_{1,1} & b_{1,2} & \dots \\
    \vdots & \ddots & \\
    b_{2n,1} &        & b_{2n,m} 
    \end{bmatrix}
    \begin{bmatrix}
    \omega_1\\
    \vdots\\
    \omega_m\\
    \end{bmatrix}
\end{aligned}
    \label{eq:mimic_dynamics}
\end{equation}
where $y_{1,j}, ~j=1, \ldots, n$ denotes the movement from each element of the computational structure, $a_{j,k}$ and $b_{j,k}$ are determined by the physical properties, and $\omega_j,~j=1, \ldots, m$ is the regulator's input. 

By inspecting Eq.~\ref{eq:mimic_dynamics}, it can be observed that the contribution of the input term $u$ based on mode generation and regulation can be separately considered through the design of $a_{j,k}$ (i.e., structure configuration and material properties) and $b_{j,k}$ (regulator or low-power actuator placement), as discussed in \cite{sihite_actuation_2023}.

We perform temporal (i.e., $t_i,~i=1, \ldots, n, \quad 0 \leq t_i \leq t_f$) discretization of Eq.~\ref{eq:mimic_dynamics} to obtain the following system of equations
\begin{equation}
\dot{Y}_{i}(t_i)=A_{i}Y_i(t_i) + B_{i}\Omega_i(t_i), \quad i=1, \ldots, n, \quad 0 \leq t_i \leq t_f
\label{eq:disc-model}    
\end{equation}
where $Y_i =\left[y_{1,1}^\top,\dots,y_{1,n}^\top\right]^\top$ embodies $n$ spatial values of the computational structure response at i-th discrete time (i.e., posture at time $t_i$). And, $\Omega_i=\left[\omega_1,\dots,\omega_m\right]$ embodies $m$ regulators actions at i-th discrete time. $A_i$ and $B_i$ are the matrices shown in Eq.~\ref{eq:mimic_dynamics} with their entries. 

We stack all of the postures and low-power inputs from the regulators from each i-th sample time, i.e., $Y_i$ and $\Omega_i$, in the vectors $Y = \left[Y^\top_1(t_1), \ldots, Y_n^\top(t_n)\right]^\top$ and $\Omega = \left[\Omega^\top_1(t_1), \ldots, \Omega^\top_m(t_n)\right]^\top$. 

We consider 2$n$ boundary conditions at the boundaries of $n$ structure elements (2 equations at each boundary) to enforce the continuity of the computational structure, given by
\begin{equation}
    r_i\left(Y(0), Y\left(t_f\right), t_f\right)=0, \quad i=1, \ldots, 2 n
\end{equation}
Since we have $m$ regulators, we consider $m$ inequality constraints given by
\begin{equation}
    g_i(Y(t_i), \Omega(t_i), t_i) \geq 0, \quad i=1, \ldots, m, \quad 0 \leq t_i \leq t_f
\end{equation}
to limit the actuation stroke from the low-power actuators.  

To approximate nonlinear dynamics from the computational structure, we employ a method based on polynomial interpolations. This method extremely simplifies the computation efforts. Consider the $n$ time intervals during a gait cycle of the dynamic morphing system, as defined previously and given by 
\begin{equation}
0=t_1<t_2<\ldots<t_n=t_f
    % \label{eq:?}
\end{equation}
We stack the states and regulator inputs $Y = \left[Y^\top_1(t_1), \ldots, Y_n^\top(t_n)\right]^\top$ and $\Omega = \left[\Omega^\top_1(t_1), \ldots, \Omega^\top_m(t_n)\right]^\top$ from the computational structure at these discrete times into a single vector denoted by $\mathcal{Y}$ and form a decision parameter vector that the optimizer finds at once. Additionally, we append the final discrete time $t_f$ as the last entry of $\mathcal{Y}$ so that gaitcycle time too is determined by the optimizer.
\begin{equation}
\mathcal{Y}=\left[Y^\top_1(t_1), \ldots, Y_n^\top(t_n), \Omega^\top_1(t_1), \ldots, \Omega^\top_m(t_n), t_f\right]^\top
\end{equation}
We approximate the regulator's action at time $t_i \leq t<t_{i+1}$ as the linear interpolation function $\tilde\Omega(t)$ between $\Omega_i(t_i)$ and $\Omega_{i+1}(t_{i+1})$ given by 
\begin{equation}
\tilde\Omega(t)=\Omega_i\left(t_i\right)+\frac{t-t_i}{t_{i+1}-t_i}\left(\Omega_{i+1}\left(t_{i+1}\right)-\Omega_i\left(t_i\right)\right)
    % \label{}
\end{equation}
We interpolate the computational structure states $Y_i(t_i)$ and $Y_{i+1}(t_{i+1})$ as well. However, we use a nonlinear cubic interpolation, which is continuously differentiable with $\dot{\tilde Y}(s)=\bm F(Y(s), \Omega(s), s)$ at $s=t_i$ and $s=t_{i+1}$, where $\bm F$ denotes the full-dynamics of Aerobat. 
% formed from the dynamics defined in /eqref{eq:ss-rep-fulldyn}

To obtain $\tilde Y(t)$, we formulate the following system of equations:
\begin{equation}
    \begin{aligned}
\tilde Y(t) &=\sum_{k=0}^3 c_k^j\left(\frac{t-t_j}{h_j}\right)^k, \quad t_j \leq t<t_{j+1}, \\
c_0^j &=Y\left(t_j\right), \\
c_1^j &=h_j \bm F_j, \\
c_2^j &=-3 Y\left(t_j\right)-2 h_j \bm F_j+3 Y\left(t_{j+1}\right)-h_j \bm F_{j+1}, \\
c_3^j &=2 Y\left(t_j\right)+h_j \bm F_j-2 Y\left(t_{j+1}\right)+h_j \bm F_{j+1}, \\
\text { where } \bm F_j &:=\bm F\left(Y\left(t_j\right), \Omega\left(t_j\right), t_j\right), \quad h_j:=t_{j+1}-t_j .
\end{aligned}
\label{eq:cubic-lobatto}
\end{equation}
The interpolation function $\tilde Y$ utilized for $Y$ needs to fulfill the computational structure's derivative requirements at discrete points and at the midpoint of sample times. By examining Eq.~\ref{eq:cubic-lobatto}, it is evident that the derivative terms at the boundaries $t_{i}$ and $t_{i+1}$ are satisfied. Hence, the only remaining constraints in the nonlinear programming problem are the collocation constraints at the midpoint of $t_i-t_{i+1}$ time intervals, the inequality constraints at $t_i$, and the constraints at $t_1$ and $t_f$, all of which are included in the optimization process.

Given that the computational structure is spatially discrete and incurs significant costs associated with its curse of dimensionality, this collocation scheme reduces the number of parameters for interpolation polynomials, thereby enhancing computational performance. We address this optimization problem using MATLAB's fmincon function.  

\section{Results and Discussions}
\label{sec:results}

Banking turn experiments were carried out with the assistance of a Kinova robotic arm, during which force and moment measurements were captured using an ATI load cell, as detailed in Figure 2. Throughout these experiments, Aerobat-$\beta$ executed flapping and banking maneuvers under varied conditions. We collected seven distinct datasets, maintaining a constant pitch angle of -15 degrees across all experiments. In four of these tests, the roll angle of Aerobat-$\beta$ was set at 10, 15, 20, and 25 degrees, respectively. For the remaining three tests, while keeping the roll angle fixed at 15 degrees, we varied the Aerobat-$\beta$'s forward speed through 0.7 m/s, 0.8 m/s, and 0.9 m/s to observe the effects on banking performance. \\
For the banking turn simulation, Aerobat’s nonlinear dynamical system was numerically solved using the fourth-order Runge-Kutta method. A collocation-based optimization controller was integrated to establish the desired roll and pitch angles. Computational efficiency was enhanced by adopting a 5-step prediction horizon. The simulation ran with a time step of 0.0001 seconds, while the controller's update frequency was set to 200 Hz, corresponding to a time step of 0.005 seconds. In the simulation parameters, Aerobat maintained a flapping rate of 3.5 Hz, matching the experimental setup. The reference pitch angle was consistently held at a negative 15 degrees, while the reference roll angle was set at 0 degrees initially for one second and then adjusted to 15 degrees for the remaining 4 seconds to execute the banking turn.

The plots in  Fig.~\ref{fig:force-moment} provide a comparative analysis of experimental and simulated data for force and moment measurements across 0.85 seconds, corresponding to three flapping gait cycles of the Aerobat. In the force graphs (Fx, Fy, Fz), the shaded regions represent the range of all experimental data, capturing the maximum and minimum values for each timestamp, which are depicted in red. The simulation data, shown in black, aligns closely with the average trend of the experimental data but exhibits some variations, especially in the Fx and Fy components. As for the moment graphs (Mx, My, Mz), both experimental and simulated data follow similar patterns. Both the forces (Fx, Fy, Fz) and moments (Mx, My, Mz) exhibit periodicity with similar magnitudes at the completion of each gait cycle, reflecting the robot’s flapping motion. The force in the z-axis ranges from -0.5 to 0.5 N in the experimental data, suggesting the flapping-induced forces are significant, while it ranges from -0.3 N to 0.3 N in simulation. The periodic positive moment along the Y-axis for both experiments and simulation suggests that the robot experiences an upward and downward pitching motion. This behavior is anticipated, considering the tail-less design of the Aerobat.

Wake structures and vorticity around the wingtip are depicted in Fig.~\ref{fig:wakes}, highlighting the dynamic interactions between the Aerobat and the air during various phases of a flapping cycle. The vorticity noticeably increases during the downstroke, signaling enhanced aerodynamic forces, and reduces as the wings are folded in the upstroke, indicating a drop in aerodynamic engagement. Initially, the vorticity distribution appears symmetrical, indicative of stable flight. Yet, with the onset of banking, an asymmetrical pattern of vorticity develops, showcasing distinct disparities across the Aerobat's wings.

Figure~\ref{fig:states} showcases the robot's dynamical states, including orientation and body angles. The pitch angle of the robot oscillates around a constant value of -15 degrees, with an amplitude ranging from 7 to 10 degrees due to the wing flapping motion. Additionally, the roll angle is precisely held at 15 degrees, enabling a significant change in the yaw angle that facilitates a smooth banking turn. Throughout the 5-second trajectory, the robot covers a distance of 4.1 meters in the x-direction and 5 meters in the y-direction. Successful 3D banking turns, achieved in both simulation and experiment, are illustrated through snapshots of Aerobat, as shown in Fig.~\ref{fig:test-setup} and Fig.~\ref{fig:sim-stick-diagram}.

% \begin{figure}
%     \centering
%     \includegraphics[width=0.9\linewidth]{example-image-a}
%     \caption{Snapshots of arm and Aerobat during the constrained banking turn at different sample times are overlaid and illustrated in this image.}
%     \label{fig:bank-turn-arm}
% \end{figure}

\section{Concluding Remarks}
\label{conclusion}

This study has successfully enhanced the accuracy of an aerodynamic Fluid-Structure Interaction (FSI) model, enabling the controlled 3D flight path tracking of the dynamic, morphing-winged drone, Aerobat. Through experimentation and simulation, including the fine-tuning of the unsteady model for banking turn maneuvers and the integration of a collocation-based control strategy, this work has demonstrated improvements in predicting complex 3D maneuvers and achieving precise flight control. 

Moving forward, future efforts will concentrate on enhancing the precision of the model, investigating how changing trajectories of flight impact Aerobat's FSI, and broadening the drone's agility to encompass a wider range of intricate aerial maneuvers. These efforts aim to bridge the gap between theoretical models and real-world applicability.

\printbibliography

\end{document}